\documentclass[final]{siamltex}
\usepackage{graphicx} 
\usepackage{mathtools}
\usepackage{latexsym}
\usepackage[mathscr]{euscript}
\usepackage{xcolor}
\usepackage{algorithm}
\usepackage{fancyhdr}
\usepackage{multirow}
\usepackage{epsfig}
\usepackage{mathptmx}       
\usepackage[most]{tcolorbox}
\usepackage{helvet}         
\usepackage{courier}        
\usepackage{type1cm}        
\usepackage{imakeidx}         
\usepackage[a4paper,textheight=23cm,textwidth=15cm]{geometry}
\usepackage{algpseudocode}
\usepackage{hyperref}
\usepackage{enumitem}
\usepackage{titlesec}
\usepackage{amsmath,amssymb,amsfonts}
\usepackage{hyperref}
\usepackage{tikz}
\usepackage{bm}
\usepackage{booktabs}
\newtheorem{remark}{remark}
\newcommand{\R}{\mathbb{R}}

\newcommand{\nd}{\noindent}

\title{Low-Rank Tensor Approximation of Weights in Large Language Models via Cosine Lanczos Bidiagonalization}
\author{A. Elichi\thanks{Université du Littoral Cote d'Opale, LMPA, 50 rue F. Buisson, 62228 Calais-Cedex, France.} \and  K. Jbilou\footnotemark[1]} 
\date{}

\begin{document}

\maketitle

\begin{abstract}
Large Language Models (LLMs) have demonstrated remarkable capabilities across diverse natural language tasks but suffer from extremely large memory footprints and computational costs. In this paper, we introduce a  tensor compression framework based on  the  c-product for computing  low-rank approximation  In the first part of our approach, we leverage the algebraic structure of the c-product to represent weight tensors—such as those in embedding layers, attention projections, and feed-forward networks—in a transform domain where frontal slices can be jointly approximated by low-rank tensor factors. This enables computationally efficient compression that exploits multi-dimensional correlations beyond traditional SVD methods. 
\end{abstract}

\section{Introduction}

The rapid development of Large Language Models (LLMs) and Transformer-based architectures
has led to unprecedented performance across a wide range of natural language processing
tasks, including language modeling, translation, and reasoning
\cite{Achiam2023,Brown2020,Touvron2023,Vaswani2017}.
However, these gains come at the cost of extremely large model sizes, resulting in high
memory footprints, significant computational demands, and limited deployability on
resource-constrained platforms.
Consequently, the design of efficient compression and approximation techniques has become
a central research challenge in modern deep learning.

Low-rank methods constitute a principled approach for model compression, motivated by the
observation that many weight matrices in deep networks exhibit substantial redundancy.
Recent works have demonstrated that rank reduction can be used to compress Transformer
models while preserving performance
\cite{Ash2023,Luo2024,Sharma2023,Yu2020}.
In particular, layer-selective strategies such as  Layer-Selective Rank Reduction (LASER) \cite{Ash2023,Xia2024} focus on reducing redundancy only
in carefully chosen layers, leading to favorable trade-offs between compression and
accuracy.
These approaches naturally motivate the use of structured matrix and tensor
decompositions that allow fine-grained control over rank and approximation error.

Tensors provide a natural mathematical framework for representing and analyzing the
multi-dimensional structure inherent in modern neural networks.
Beyond classical matrix-based methods, tensor decompositions such as CANDECOMP/PARAFAC,
Tucker, and Tensor Train (TT) offer powerful tools for capturing higher-order correlations
in data and model parameters \cite{Hitchcock1927,Kolda2009,Oseledets2011,Tucker1966}.
In the context of deep learning, tensor methods have been successfully applied to neural
network compression, embedding representations, and efficient inference
\cite{Newman2024,Xu2023}.
Among these, the Tensor Train format stands out due to its favorable algebraic properties,
numerical stability, and linear storage complexity with respect to tensor order.

Parallel to these developments, tensor–tensor products based on invertible linear
transforms have emerged as a powerful extension of classical matrix algebra.
The $\mathcal L$-product framework introduced in \cite{Aeron2015} unifies several tensor
products by applying a linear transform along one mode, thereby enabling efficient and
theoretically sound tensor computations.
Special instances of this framework include the t-product based on the FFT
\cite{Elhachimi2024,Kilmer2011} and the c-product based on the discrete cosine transform \cite{Hached2021},
which enjoys real-valued arithmetic and strong energy compaction properties.
These products admit tensor analogues of fundamental matrix concepts, such as orthogonality,
singular value decomposition, and rank, enabling the development of low-rank tensor
approximations and scalable numerical algorithms.

A key computational challenge in tensor-based methods is the efficient extraction of
dominant spectral information from large-scale tensors.
Tensor Lanczos bidiagonalization under the $\mathcal L$-product has proven to be an
effective tool for approximating a small number of extreme singular triplets without forming full tensor decompositions \cite{Dufrenois2023,Elhachimi2024,Hached2021}.
Such Krylov-based techniques are particularly well suited for large models, where only a
low-rank approximation of the weights is required.
They provide the numerical foundation for scalable rank reduction strategies in
high-dimensional settings.

In this work, we bring together these ideas to develop a unified tensor-based framework for
efficient compression of large neural models.
We leverage the c-product and its associated c-SVD to perform structured, transform-based
low-rank approximations of weight tensors, and we employ tensor Lanczos bidiagonalization
to compute only the most significant spectral components.
Building on these tools, we introduce Tensor Layer-Selective Rank Reduction (TLASER), which
extends LASER-style rank reduction to the tensor setting under the $\mathcal L$-product.
Furthermore, for high-order tensor representations arising from embedding vectors and other
model components, we exploit the Tensor Train decomposition to obtain compact and stable
factorizations with controllable approximation error.

The remainder of this paper is organized as follows.
Section~\ref{preliminaries} introduces basic tensor notation, matricization, and the
$n$-mode product. Section~2 presents the $\mathcal L$-product framework, with a particular focus on the c-product and its associated spectral decomposition.
Section~\ref{sec 3} describes tensor Lanczos bidiagonalization under the c-product and its use in computing approximate singular triplets.
Tensor Layer-Selective Rank Reduction is introduced thereafter, followed by a detailed presentation of the Tensor Train decomposition for high-order tensor problems. The last section is devoted to some experimental results.

\section{Preliminaries and Notation}
\label{preliminaries}

\subsection{Tensors and Basic Notation}

A \emph{tensor} is a multidimensional array of numerical values, where the number of indices (or modes) determines its order. Scalars, vectors, and matrices correspond to tensors of order 0, 1, and 2, respectively. Higher-order tensors (order $\ge 3$) are denoted using calligraphic letters.  

\begin{itemize}
    \item The \emph{order} (or mode) of a tensor $\mathcal{A} \in \mathbb{R}^{n_1 \times n_2 \times \cdots \times n_N}$ is $N$.
    \item Each entry of $\mathcal{A}$ is denoted by $a_{i_1, \dots, i_N}$, where $1 \le i_j \le n_j$, $j=1, \dots, N$. That is,
    \[
        \mathcal{A}_{i_1 \dots i_N} = a_{i_1, \dots, i_N}.
    \]
    \item The Frobenius norm of $\mathcal{A}$ is defined as
    \[
        \|\mathcal{A}\|_F^2 = \sum_{i_1=1}^{n_1}\sum_{i_2=1}^{n_2} \dots \sum_{i_N=1}^{n_N} a_{i_1\dots i_N}^2.
    \]
\end{itemize}

A \emph{slice} of a tensor is obtained by fixing all indices except two, resulting in a matrix. A \emph{fiber} is obtained by fixing all indices except one; in a third-order tensor $\mathcal{A} \in \mathbb{R}^{n_1 \times n_2 \times n_3}$, the mode-3 fibers (also called tubes) are obtained by fixing the first two indices.  

\subsection{Matricization and $n$-Mode Product}

\emph{Matricization} (or unfolding) reshapes a tensor into a matrix. For an $N$-mode tensor $\mathcal{A}$, its $n$-mode unfolding is denoted $\mathcal{A}_{(n)}$. The $n$-mode product of a tensor with a matrix $X \in \mathbb{R}^{J \times I_n}$ is defined as follows:

\begin{definition}[$n$-Mode Product]
Given $\mathcal{A} \in \mathbb{R}^{I_1 \times \cdots \times I_N}$, the $n$-mode product $\mathcal{A} \times_n X \in \mathbb{R}^{I_1 \times \cdots \times I_{n-1} \times J \times I_{n+1} \times \cdots \times I_N}$ has entries
\[
(\mathcal{A} \times_n X)_{i_1 \dots i_{n-1} j i_{n+1} \dots i_N} = \sum_{i_n=1}^{I_n} a_{i_1\dots i_N} x_{j i_n}.
\]
Equivalently, in matricized form:
\[
\mathcal{B} = \mathcal{A} \times_n X \iff \mathcal{B}_{(n)} = X \mathcal{A}_{(n)}.
\]
\end{definition}

\nd When $X$ is a vector $v \in \mathbb{R}^{I_n}$, the $n$-mode product produces a tensor of order $N-1$:
\[
(\mathcal{A} \bar\times_n v)_{i_1 \dots i_{n-1} i_{n+1} \dots i_N} = \sum_{i_n=1}^{I_n} a_{i_1 \dots i_N} v_{i_n}.
\]

\nd Classical tensor decompositions such as CANDECOMP/PARAFAC (CP) \cite{Hitchcock1927}, Tucker \cite{Tucker1966}, and Tensor Train (TT) \cite{Oseledets2011} utilize the $n$-mode product.

\section{The $\mathcal{L}$-Product}

The $\mathcal{L}$-product \cite{Aeron2015} is a general tensor multiplication framework that unifies several tensor products via a linear transform along the third mode.

\begin{definition}[$\mathcal{L}$-Transform]
Let $Z \in \mathbb{R}^{p \times p}$ be invertible. For $\mathcal{A} \in \mathbb{R}^{n \times m \times p}$, define
\[
\mathcal{L}(\mathcal{A}) = \mathcal{A} \times_3 Z, \quad
\mathcal{L}^{-1}(\mathcal{A}) = \mathcal{A} \times_3 Z^{-1}.
\]
\end{definition}

\medskip 
Denote the transformed tensor by $\widehat{\mathcal{A}} = \mathcal{L}(\mathcal{A})$, with frontal slices $\widehat{A}_i = \widehat{\mathcal{A}}(:,:,i)$.  

\medskip 
\begin{definition}[Facewise Product]
For $\mathcal{A} \in \mathbb{R}^{m \times l \times p}$ and $\mathcal{B} \in \mathbb{R}^{l \times n \times p}$, the facewise product $\mathcal{C} = \mathcal{A} \triangle \mathcal{B}$ is defined slice-wise as
\[
\widehat{C}_i = \widehat{A}_i \widehat{B}_i, \quad i=1,\dots,p.
\]
\end{definition}

\begin{definition}[$\cal L$-identity tensor]
A tensor $\mathcal{I} \in \mathbb{R}^{m \times m \times p}$ is the
\emph{identity tensor} if
\[
\mathcal{I}(i,i,:) = \mathbf{e}, \qquad i = 1, \ldots, m,
\]
and  $\mathbf{e}$ is the ${\mathcal L}$-identity tube defined by:
\[
\mathbf{e} = {\mathcal L}^{-1}(\mathbf{1})= \mathbf{1} \times_3 Z^{-1},
\]
where $\mathbf{1} \in \mathbb{R}^{1 \times 1 \times p}$ denotes the tube containing all ones.
\end{definition}

\medskip

\begin{definition}[$\mathcal L$-transpose]
Let ${\mathcal A} \in \mathbb{R}^{m\times n \times p}$.
Then its transpose ${\mathcal B}={\mathcal A}^{T} \in \mathbb{R}^{n \times m \times p}$
is defined such that 
\[
\widehat{B}_i = {\widehat{A}_i}^{T},
\qquad i = 1, \ldots, p.
\]
\end{definition}

\medskip

\begin{definition}
A square tensor $\mathcal{Q} \in \mathbb{R}^{m \times m \times p}$ is 
\begin{itemize}
\item Orthogonal if
\[
\mathcal{Q}^T *_\mathcal{L} \mathcal{Q}
  = \mathcal{Q} *_\mathcal{L} \mathcal{Q}^T
  = \mathcal{I}_{mmp}.
\]
If $\mathcal{Q}$ is f-orthogonal, then all frontal slices 
$\widehat{Q}_i$, $i=1,\ldots,p$, of $\widehat{\mathcal{Q}} = {\mathcal L}(\mathcal{Q})$ are orthogonal matrices.
\item  f-diagonal if 
all its frontal slices are diagonal matrices.
\item  Invertible if 
there exists $\mathcal{X} \in \mathbb{R}^{m \times m \times p}$ (denoted $\mathcal{X}=\mathcal{A}^{-1})$  such that
\[
\mathcal{X} *_{\mathcal L} \mathcal{A}= \mathcal{A} *_{\mathcal L} \mathcal{X}
  = \mathcal{I}_{mmp}.
\]
\end{itemize}
\end{definition}

\medskip 

\subsubsection{The tensor cosine product} 

    
  \nd The  cosine product (c-product) denoted by $*_c$  is obtained by selecting in the expression of $\mathcal L$ the following matrix 
    \[
    Z = W^{-1}_{p} C_{p}(I_{p}+L_{p}),
    \]
    \noindent where $C_{p}$ is the discrete cosine transform matrix and $Z_{p}$ is a circulant shift matrix. In that case, the c-product uses the Discrete Cosine Transform (DCT) along the tubes:    $${\widehat {\mathcal A}}= DCT({\mathcal A}, 3).$$
   The c-product of two tensors $\mathcal{A} \in \mathbb{R}^{m \times l \times p}$ and $\mathcal{B} \in \mathbb{R}^{l \times n \times p}$ is the tensor ${\mathcal C} \in \mathbb{R}^{m \times n \times p} $ defined by  $${\mathcal C}= {\mathcal A} *_c {\mathcal B} \Longleftrightarrow {\widehat C}_i= {\widehat A}_i *_c {\widehat B}_i,\; i=1,\ldots,p,$$
    \nd Consequently, the $\mathcal{L}$-product establishes a unified algebraic framework for tensor computations based on linear transforms. The $\mathcal{L}$-product generalizes the t-product (using FFT) and c-product (using DCT) and allows low-rank tensor approximations, truncated $\mathcal{L}$-SVD, and efficient spectral computations.

    \medskip
    \begin{remark}
The DCT-based c-product is real-valued and achieves superior energy compaction, which is particularly useful for compression, denoising, and low-rank approximations.
\end{remark}
\nd     In what follows, we consider only the c-product denoted by $*_c$.

\medskip


\begin{definition}[c-SVD]
Any $\mathcal{A} \in \mathbb{R}^{m \times n \times p}$ admits a c-SVD decomposition as follows
\begin{equation}\label{svd1}
\mathcal{A} = \mathcal{U} *_c \mathcal{S} *_c \mathcal{V}^T,
\end{equation}
where $\mathcal{U}$ and $\mathcal{V}$ are $\mathcal{L}$-orthogonal tensors, and $\mathcal{S}$ is f-diagonal. The norms of the tubes of $\mathcal{S}$ are called singular values \cite{Elhachimi2024,Kilmer2011}.
\end{definition}

\nd \nd The decomposition \eqref{svd1} allows $\mathcal{A}$ to be expressed as
\[
\mathcal{A}=\sum_{i=1}^{\min(m,n)}\vec{\mathcal{U}}_i*\bm{s}_{i}*_c\vec{\mathcal{V}}^H_i 
= \sum_{i=1}^{\min(m,n)}\bm{s}_{i}*_c\vec{\mathcal{U}}_i*_c\vec{\mathcal{V}}^H_i,
\]
where $\bm{s}_i = \mathcal{S}(i,i,:)$ represents the $i^{th}$ singular tube, and 
$\vec{\mathcal{U}}_i = \mathcal{U}(:,i,:)$ and $\vec{\mathcal{V}}_i = \mathcal{V}(:,i,:)$ 
are left and the right singular slices of $\mathcal{A}$, respectively. The singular
tubes are ordered according to decreasing norm, i.e.,
\[
\|\bm{s}_1\|\ge\|\bm{s}_2\|\ge\ldots\ge\|\bm{s}_{\min(m,n)}\|\geq 0,
\]
where $\|\cdot\|$ denotes the Euclidean vector norm. We refer to
$\{\bm{s}_i,\vec{\mathcal{U}}_i,\vec{\mathcal{V}}_i\}$ as the $i$th singular triplet of
$\mathcal{A}$. The large singular triplets are associated with singular tubes of 
(relatively) large magnitude, while small singular triplets are associated with singular 
tubes of (relatively) small magnitude. Algorithm \ref{alg:t-svd} outlines the computation 
of the c-svd of a third-order tensor.

\begin{algorithm}
    \caption{The c-SVD}
    \label{alg:t-svd}
    \textbf{Input:} $\mathcal{A} \in \mathbb{\R}^{m \times n \times p}$\\
    \textbf{Output:} $\mathcal{U} \in \mathbb{\R}^{m \times n \times p}$, $\mathcal{S} \in \mathbb{\R}^{m \times n \times p}$, $\mathcal{V} \in \mathbb{\R}^{m \times n \times p}$\\
   \begin{itemize}
       \item  Compute $\widehat{\mathcal{A}}= DCT({\mathcal A},3)$.\\
       \item     For  {$i = 1, \ldots, p$}\\
               $\left[\widehat{{U}}_{i}, \widehat{{S}}_{i}, \widehat{{V}}_{i}\right] = \texttt{svd}\left(\widehat{{A}}_{i}\right)$,  where $\widehat{{A}}_{i}$ is the $i$-th frontal slice of $\widehat{\mathcal{A}}$\\
            EndFor\\
      \item   Return $\mathcal{U}$, $\mathcal{S}$, and $\mathcal{V}$ by the inverse of DCT.
   \end{itemize}
\end{algorithm}

\medskip
Algorithm \ref{alg:t-svd} computes the c-Singular Value Decomposition (c-SVD) of a third-order tensor under the c-product framework. The c-SVD generalizes the classical matrix SVD to tensors by exploiting a linear transform along the third mode, which allows the tensor decomposition to be performed slice-wise in the transform domain.
\\
The algorithm first applies the c-transform (e.g., Discrete Cosine Transform) along the third mode of the input tensor $\mathcal{A}$, producing $\widehat{\mathcal{A}}$. In this domain, the c-product reduces to standard matrix multiplication on each frontal slice. For each slice $\widehat{{A}}_{i}$, the classical matrix SVD is computed:
\[
\widehat{{A}}_{i} = \widehat{{U}}_{i} \, \widehat{{S}}_{i} \, \widehat{{V}}_{i}^T,\; i=1,\ldots,p.
\]
The slice-wise SVDs capture the spectral information of the tensor in the transform domain.
After computing all slice-wise SVDs, the inverse $\mathcal{L}$-transform is applied to assemble the tensor factors $\mathcal{U}$, $\mathcal{S}$, and $\mathcal{V}$ in the original domain. The resulting tensors satisfy the c-SVD factorization
\[
\mathcal{A} = \mathcal{U} *_c \mathcal{S} *_c \mathcal{V}^T,
\]
where $\mathcal{U}$ and $\mathcal{V}$ are $\mathcal{L}$-orthogonal and $\mathcal{S}$ is f-diagonal.
\\
Algorithm \ref{alg:t-svd} is conceptually simple and numerically robust since it reduces the tensor problem to standard matrix SVDs. It is also naturally parallelizable, as the slice-wise SVDs are independent. However, computing the full c-SVD can be costly for large tensors, and in practice, partial methods such as tensor Lanczos bidiagonalization are often used to compute only dominant singular triplets efficiently. Overall, this algorithm provides a foundation for low-rank tensor approximations, spectral analysis, and tensor-based model compression within the c-product framework.

\medskip 

%
\noindent
The classical notion of matrix rank can be extended in a natural and meaningful way
to third-order tensors within the $\mathcal{L}$-product framework. Since tensor
algebra under transform-based products reduces many operations to matrix computations
in the transform domain, multiple rank notions arise depending on how the spectral
information across modes is aggregated. In this work, we consider two widely used
definitions of tensor rank under the $c$-product, each capturing different aspects of
the multilinear structure.

\begin{definition}[Average rank under the $c$-product]
\index{c-average-rank}
Let $\mathcal{A} \in \mathbb{R}^{m \times n \times p}$ be a third-order tensor and let
$\widehat{\mathcal{A}} = \mathcal{L}(\mathcal{A})$ denote its representation in the
transform domain. The \emph{$c$-average rank} of $\mathcal{A}$ is defined as
\[
\text{Rank}_a(\mathcal{A})
=
\frac{1}{p}
\sum_{i=1}^{p}
\text{rank}\!\left(\widehat{\mathcal{A}}^{(i)}\right),
\]
where $\widehat{\mathcal{A}}^{(i)}$ denotes the $i$-th frontal slice of
$\widehat{\mathcal{A}}$. This quantity measures the average matrix rank across all
transformed frontal slices and provides a global indicator of the overall spectral
complexity of the tensor.
\end{definition}

\medskip

\begin{definition}[Tubal rank under the $c$-product]
\index{c-tubal-rank}
Let $\mathcal{A} \in \mathbb{R}^{m \times n \times p}$ and consider its $c$-SVD
\[
\mathcal{A}
=
\mathcal{U} *_c
\mathcal{S} *_c
\mathcal{V}^\top,
\]
where $\mathcal{S}$ is an $f$-diagonal tensor. The \emph{$c$-tubal rank} of $\mathcal{A}$
is defined as
\begin{equation}
\label{eq:L-tubal-rank}
\text{Rank}_T(\mathcal{A})
=
\text{card}
\left\{
i \;\middle|\;
\exists\, j \in \{1,\dots,p\}
\text{ such that }
\mathcal{S}(i,i,j) \neq 0
\right\}.
\end{equation}
Equivalently, the tubal rank counts the number of nonzero diagonal tubes of the
$f$-diagonal tensor $\mathcal{S}$, and thus quantifies the intrinsic dimensionality
of $\mathcal{A}$ along the transform mode.
\end{definition}

\medskip

\noindent
While the $c$-average rank provides a useful measure of global spectral richness,
it does not directly correspond to a minimal representation of the tensor. In contrast,
the $c$-tubal rank plays a role analogous to the matrix rank in classical linear algebra:
it determines the smallest number of rank-one tensor components required to represent
$\mathcal{A}$ under the $c$-product. For this reason, and due to its close connection
with low-rank tensor approximation and truncation via the $c$-SVD, the tubal rank is
adopted as the notion of tensor rank throughout this paper.

\section{Tensor Lanczos bidiagonalization under the c-product}
\label{sec 3}
The generalization of matrix decompositions to higher-order tensors through the c-product has proven useful in multi-dimensional data analysis. In particular, an application to face recognition using a database of color facial expression images is presented in \cite{Elhachimi2023}. The proposed numerical approach relies on computing a small number of the dominant singular triplets of a large third-order tensor, which is accomplished via tensor Lanczos bidiagonalization.
\\
Let $\mathcal{A} \in \mathbb{R}^{m \times n\times p}$ be a third-order tensor. The 
application of $k\ll\min\{m, n, p\}$ steps of the tensor Lanczos bidiagonalization process 
generically yields the tensors
\begin{equation}\label{tenPQ}
\mathcal{P}_k = \left[\vec{\mathcal{P}}_1, \ldots, \vec{\mathcal{P}}_k\right] \in 
\mathbb{R}^{m \times n\times p}, \quad 
\mathcal{Q}_k = \left[\vec{\mathcal{Q}}_1, \ldots, \vec{\mathcal{Q}}_k \right] \in 
\mathbb{R}^{m \times n\times p}.
\end{equation}
The lateral slices of $\mathcal{P}_k$ and $\mathcal{Q}_k$ form bases for the tensor Krylov
subspaces
\[
\begin{split}
\bm{K}_k(\mathcal{A}^H * \mathcal{A}, \vec{\mathcal{P}}_1) = 
\text{span}\left\{ \vec{\mathcal{P}}_1, \mathcal{A}^H * \mathcal{A} *_c \vec{\mathcal{P}}_1,
\ldots, (\mathcal{A}^H *_c \mathcal{A})^{k-1} * \vec{\mathcal{P}}_1 \right\},\\
\bm{K}_k(\mathcal{A} *_c  \mathcal{A}^H, \vec{\mathcal{Q}}_1) = 
\text{span}\left\{ \vec{\mathcal{Q}}_1, \mathcal{A} * \mathcal{A}^H *_c \vec{\mathcal{Q}}_1,
\ldots, (\mathcal{A} *_c \mathcal{A}^H)^{m-1} * \vec{\mathcal{Q}}_1 \right\},
\end{split}
\]
respectively.\\

Algorithm \ref{alg:t_product_lanczos} describes the computations of the tensors involved in tensor Lanczos process.

\begin{algorithm}[H]
\caption{Partial tensor Lanczos bidiagonalization with the c-product}
\label{alg:t_product_lanczos}

\textbf{Input:} 
$\mathcal{A} \in \mathbb{R}^{m \times n \times p}$, initial unit-norm slice 
$\vec{\mathcal{P}}_1 \in \mathbb{R}^{n \times 1 \times p}$, number of steps $k \in \mathbb{N}^*$.

\textbf{Output:} 
Tensors $\mathcal{P}_k \in \mathbb{R}^{m \times n \times p}$ and 
$\mathcal{Q}_k \in \mathbb{R}^{m \times n \times p}$ with f-orthonormal lateral slices, 
an f-upper bidiagonal tensor $\mathcal{B}_k \in \mathbb{R}^{m \times n \times p}$, 
and a residual slice $\vec{\mathcal{R}}_k \in \mathbb{R}^{n \times 1 \times p}$.
\medskip
\begin{enumerate}
    \item Set $\mathcal{P}_1 = [\vec{\mathcal{P}}_1]$.
    \item Compute $\vec{\mathcal{Q}}_1 = \mathcal{A} *_c \vec{\mathcal{P}}_1$.
    \item Normalize $\vec{\mathcal{Q}}_1$ to obtain 
    $\left[\vec{\mathcal{Q}}_1, \bm{\alpha}_1\right] = \text{Normalize}(\vec{\mathcal{Q}}_1)$.
    \item Set $\mathcal{Q}_1 = [\vec{\mathcal{Q}}_1]$ and 
    $\mathcal{B}_k(1,1,:) = \bm{\alpha}_1$.

    \item For $i = 1, \ldots, k$:
    \begin{enumerate}
        \item Compute the residual
        \[
        \vec{\mathcal{R}}_i = \mathcal{A}^H *_c \vec{\mathcal{Q}}_i - \bm{\alpha}_i *_c \vec{\mathcal{P}}_i .
        \]
        \item Reorthogonalize
        \[
        \vec{\mathcal{R}}_i = \vec{\mathcal{R}}_i - \mathcal{P}_i * 
        (\mathcal{P}_i^H *_c \vec{\mathcal{R}}_i).
        \]

        \item If $i < k$:
        \begin{enumerate}
            \item Normalize the residual
            \[
            \left[\vec{\mathcal{P}}_{i+1}, \bm{\beta}_i\right] =
            \text{Normalize}(\vec{\mathcal{R}}_i).
            \]
            \item Expand the basis 
            $\mathcal{P}_{i+1} = [\mathcal{P}_i, \vec{\mathcal{P}}_{i+1}]$ and set
            $\mathcal{B}_k(i,i+1,:) = \bm{\beta}_i$.
            \item Compute
            \[
            \vec{\mathcal{Q}}_{i+1} =
            \mathcal{A} *_c \vec{\mathcal{P}}_{i+1}
            - \bm{\beta}_i *_c \vec{\mathcal{Q}}_i .
            \]
            \item Reorthogonalize
            \[
            \vec{\mathcal{Q}}_{i+1} =
            \vec{\mathcal{Q}}_{i+1}
            - \mathcal{Q}_i *_c (\mathcal{Q}_i^H * \vec{\mathcal{Q}}_{i+1}).
            \]
            \item Normalize
            \[
            \left[\vec{\mathcal{Q}}_{i+1}, \bm{\alpha}_{i+1}\right]
            = \text{Normalize}(\vec{\mathcal{Q}}_{i+1}).
            \]
            \item Expand the basis 
            $\mathcal{Q}_{i+1} = [\mathcal{Q}_i, \vec{\mathcal{Q}}_{i+1}]$ and set
            $\mathcal{B}_k(i+1,i+1,:) = \bm{\alpha}_{i+1}$.
        \end{enumerate}
    \end{enumerate}
    \item End
\end{enumerate}
\end{algorithm}

Algorithm \ref{alg:t_product_lanczos} is a tensor generalization of the classical Golub–Kahan Lanczos bidiagonalization method, adapted to third-order tensors under the c-product framework. Its main objective is to compute low-dimensional tensor Krylov subspaces that capture the dominant spectral information of the input tensor $\mathcal{A}$ without explicitly forming a full c-SVD.
\\
The algorithm constructs two sequences of lateral slices, collected in the tensors $\mathcal{P}_k$ and $\mathcal{Q}_k$, whose columns form f-orthonormal bases of tensor Krylov subspaces associated with $\mathcal{A}^H *_c \mathcal{A}$ and $\mathcal{A} *_c \mathcal{A}^H$, respectively. These bases play the role of left and right Lanczos vectors in the matrix setting.
\\
At each iteration, the algorithm alternates between applying $\mathcal{A}$ and its c-adjoint $\mathcal{A}^H$ to the current basis vectors. The coefficients $\bm{\alpha}_i$ and $\bm{\beta}_i$ are tube scalars that populate the diagonal and superdiagonal of the f-upper bidiagonal tensor $\mathcal{B}_k$. This tensor is the compact representation of $\mathcal{A}$ projected onto the generated Krylov subspaces and serves as a low-dimensional surrogate for spectral computations.
\\
Reorthogonalization steps are explicitly included to maintain f-orthogonality of the generated lateral slices. This is essential for numerical stability, as loss of orthogonality may otherwise occur due to finite-precision arithmetic, especially when computing several Lanczos steps or when the tensor exhibits clustered singular values.
\\
The normalization operations ensure that each newly generated slice has unit Frobenius norm, mirroring the normalization of vectors in the classical Lanczos process. The residual slice $\vec{\mathcal{R}}_i$ measures the component of $\mathcal{A}^H *_c \vec{\mathcal{Q}}_i$ orthogonal to the current Krylov subspace and provides a stopping criterion when its norm becomes sufficiently small. 
After $k$ iterations, the algorithm yields a small bidiagonal tensor $\mathcal{B}_k$ whose c-SVD approximates the leading singular triplets of the original tensor $\mathcal{A}$. These approximations can be lifted back to the original tensor space using the computed bases $\mathcal{P}_k$ and $\mathcal{Q}_k$, resulting in accurate low-rank approximations with significantly reduced computational cost.
\\
Overall, Algorithm \ref{alg:t_product_lanczos} is well suited for large-scale tensor problems where only a few dominant singular components are required. Its reliance on tensor–tensor products and matrix-level operations in the transform domain makes it particularly attractive for applications such as tensor rank reduction, denoising, and compression of large neural network weight tensors.

\medskip \nd A detailed
derivation of the algorithm can be found in \cite{Elhachimi2024,Hached2021}. Algorithm \ref{alg:t_product_lanczos} is based on the following relations:  
\begin{eqnarray}
\mathcal{A} *_c \mathcal{P}_k &=& \mathcal{Q}_k * \mathcal{B}_k \label{eq 2} \\
\mathcal{A}^H *_c \mathcal{Q}_k &=& \mathcal{P}_k * \mathcal{B}_k^H + 
`\vec{\mathcal{R}}_k *_c \vec{\mathcal{E}}_k^H,  \label{eq 3}
\end{eqnarray}
where the slice  $\vec{\mathcal{R}}_k$ is f-orthonormal to all the lateral slices of 
$\mathcal{P}_k$, and $\vec{\mathcal{E}}_k$ denotes the canonical lateral slice whose elements are zero except for the first element of the $k$th tube, which equals
1. The tensor $\mathcal{B}_k$ is upper bidiagonal with the structure
\[
\mathcal{B}_k = \begin{bmatrix}
    \bm{\alpha}_1 & \bm{\beta}_1 \\
    & \bm{\alpha}_2 & \bm{\beta}_2 \\
    & & \ddots & \ddots \\
    & & & \bm{\alpha}_{k-1} & \bm{\beta}_{k-1} \\
    & & & & \bm{\alpha}_k
\end{bmatrix} \in \mathbb{R}^{m \times m \times p}.
\]
where $\bm{\alpha}_i$ and $\bm{\beta}_i$ denote the coefficients that are determined by
the algorithm. Normalization of the remainder slice  $\vec{\mathcal{R}}_k$ in 
\eqref{eq 3} gives
\begin{equation}
\vec{\mathcal{R}}_k=\vec{\mathcal{P}}_k*_c \bm{\beta}_k,\label{eq beta}
\end{equation}
where $\vec{\mathcal{P}}_{k+1}$ is of unit norm and $\bm{\beta}_k$ is a tube; see \cite{Elhachimi2024}. Eq. \ref{eq 3} can now be expressed as
\[
\mathcal{A}^H*_c \mathcal{Q}_k=\mathcal{P}_{k+1}*_c \mathcal{B}_{k,k+1}^H,
\]
where the tensor $\mathcal{P}_{k+1}$ is given by appending the lateral slice 
$\vec{\mathcal{P}}_{k+1}$ to $\mathcal{P}_k$, i.e., 
$\mathcal{P}_{k+1}=\left[\mathcal{P}_k,\vec{\mathcal{P}}_{m+1}\right]$. Moreover, the 
tensor $\mathcal{B}_{k,k+1}$ is defined by 
\begin{equation}
\mathcal{B}_{k,k+1}=\begin{bmatrix}
	\mathcal{B}_k \\
	& \bm{\beta}_{k+1}
\end{bmatrix}\in \mathbb{R}^{k\times (k+1) \times p}.\label{eq Bk}
\end{equation}

\medskip\nd Let us see now how to approximation  a few singular triplets of a large  tensor ${\mathcal A}\in {\R}^{m \times n \times p}$.  We approximate the first $l$ singular triplets (of $\mathcal{A}$):  
$$\left\{\bm{s}_{i,k}^{\mathcal{A}}, \vec{\mathcal{U}}_{i,k}^\mathcal{A}, 
\vec{\mathcal{V}}_{i,k}^\mathcal{A}\right\},\;\; i=1,\ldots,l,\;\text{with}\;  l<k.$$
by 
\begin{equation}
\bm{s}_{i,k}^\mathcal{A} = \bm{s}_i, \quad 
\vec{\mathcal{U}}_{i,k}^\mathcal{A} = \mathcal{Q}*_c  \vec{\mathcal{U}}_i, \quad \text{and}\quad 
\vec{\mathcal{V}}_{i,k}^\mathcal{A} = \mathcal{P}_k *_c \vec{\mathcal{V}}_i,\; i=1,\ldots,l \label{eq 5}
\end{equation}
where $\left\{\bm{s}_i, \vec{\mathcal{U}}_i, \vec{\mathcal{V}}_i\right\}$, $i=1\ldots,k$ are the
singular triplets of $\mathcal{B}_k$ satisfying 
\[
\mathcal{B}_k * \vec{\mathcal{V}}_i = \vec{\mathcal{U}}_i * \bm{s}_i, \quad 
\mathcal{B}_k^H * \vec{\mathcal{U}}_i = \vec{\mathcal{V}}_i * \bm{s}_i.
\]
It is shown in \cite{Elhachimi2023} that the approximate singular triplets satisfy 
\[
\mathcal{A} * \vec{\mathcal{V}}_{i,m}^\mathcal{A} = \vec{\mathcal{U}}_{i,k}^\mathcal{A} *
\bm{s}_{i,k}^\mathcal{A}, \quad \text{and}\quad  \mathcal{A}^H * \vec{\mathcal{U}}_{i,k}^\mathcal{A} =
\vec{\mathcal{V}}_{i,k}^\mathcal{A} *_c \bm{s}_{i,k}^\mathcal{A} + \bm{\beta}_k *_c 
\vec{\mathcal{P}}_{k+1} * \vec{\mathcal{E}}_k^H *_c \vec{\mathcal{U}}_i.
\]
It is desirable that the residual term 
$\bm{\beta}_k *_c \vec{\mathcal{P}}_{k+1} * \vec{\mathcal{E}}_k^H * \vec{\mathcal{U}}_i$ has
small Frobenius norm. This requirement suggests that the following inequalities should be satisfied 
\begin{equation}
\left\Vert \bm{\beta}_k *_c \vec{\mathcal{P}}_{k+1} *_c \vec{\mathcal{E}}_k^H *_c 
\vec{\mathcal{U}}_i\right\Vert_F \leq \varepsilon 
\quad i=1,\ldots, k. \label{conv}
\end{equation}
\nd When these inequalities are not satisfied for a prescribed tolerance $\varepsilon$, enhancement strategies such as augmentation with Ritz lateral slices or harmonic Ritz lateral slices can be employed to refine the approximations of the singular triplets; see \cite{Elhachimi2024}. Moreover, \cite{Elhachimi2024} presents a complete GPU-accelerated Python implementation for computing a small number of extreme singular triplets of large-scale tensors using the tensor Lanczos bidiagonalization algorithm under the t-product framework, incorporating both Ritz and harmonic Ritz lateral slice augmentations. Here, we can use the same procedure when using the c-product as it is the case in our work.

\section{Tensor LAyer SElective Rank Reduction (TLASER) via the $\mathcal L$-product}

Tensor Layer-Selective Rank Reduction (TLASER) is a method aimed at enhancing the efficiency of Transformer-based models, especially in large-scale applications such as Large Language Models (LLMs). LASER achieves this by selectively reducing the rank of the weight matrices, which not only compresses the model but also maintains its performance. Unlike traditional pruning techniques, LASER applies rank reduction in a targeted manner, focusing on specific layers of the model. This selective approach improves both computational efficiency and the robustness of the model. In this section, we will delve into the workings of LASER, explore its key benefits, and discuss its practical application to Transformer models.
\\
Once the weight tensor is decomposed, TLASER performs rank reduction by discarding small singular values in \( \mathcal{S} \), which correspond to less important or noisy components. The rank-reduced approximation of the weight tensor is given by:
\[
\mathcal{W}_r = \mathcal{U}_r *_{\mathcal L} \mathcal{S}_r *_{\mathcal L}\mathcal{V}_r^T
\]
where \( \mathcal{S}_r \) retains only the largest \( r \) singular values, significantly reducing the rank of the matrix.
\\ TLASER applies rank reduction selectively to specific layers. The key idea is to target and reduce the rank of those layers that contribute the most to redundancy. This method of layer-selective rank reduction allows for greater optimization without negatively impacting the performance of the model.
\\
The rank reduction performed by TLASER acts as a form of denoising. By removing redundant and less informative components, LASER reduces the model’s parameter space and helps the model focus on the more important features. This compression leads to a lighter, faster model with lower memory requirements.\\
In Transformer models, TLASER can be applied to different components, including:
\begin{itemize}
    \item Tensor Feed-Forward Networks (FFNs): TLASER is particularly effective in reducing the rank of the weight tensors in FFN layers, which often contain a large amount of redundancy.
    \item Tensor Multi-Head Attention (MHA): The attention mechanisms of Transformers can also benefit from TLASER, especially in reducing redundant components in the query, key, and value matrices.
\end{itemize}

\medskip
\section{TLASER: Tensor Layer-Selective Rank Reduction}
\label{sec:tlaser}
We now present TLASER (Tensor LAyer-SElective Rank Reduction), a tensor extension of the LASER method \cite{Sharma2023} that exploits the natural multi-head structure of transformer weights within the c-product framework.
\subsection{Tensorization of Transformer Weights}
Transformer attention layers contain weight matrices with inherent multi-head structure. We exploit this structure through tensorization.
\begin{definition}[Multi-Head Attention Weight Structure]
\label{def:mha_structure}
For a transformer with model dimension $d_m$, $n_h$ attention heads, and head dimension $d_h = d_m/n_h$, the query projection weight $W_Q \in \mathbb{R}^{d_m \times d_m}$ is naturally organized as:
\[
W_Q = \begin{bmatrix} W_Q^{(1)} \\ W_Q^{(2)} \\ \vdots \\ W_Q^{(n_h)} \end{bmatrix}, \quad W_Q^{(h)} \in \mathbb{R}^{d_h \times d_m},
\]
where $W_Q^{(h)}$ is the query projection matrix for head $h$.
\end{definition}

\begin{definition}[Attention Tensorization Operator]
	\label{def:attention_tensorization}
	Let $W \in \mathbb{R}^{d_m \times d_m}$ be an attention weight matrix with $d_m = n_h \cdot d_h$.
	The \emph{attention tensorization operator}
	\[
	\Phi_{\textup{attn}} : \mathbb{R}^{d_m \times d_m} \longrightarrow \mathbb{R}^{d_h \times d_m \times n_h}
	\]
	associates to $W$ a third-order tensor $\mathcal{W} = \Phi_{\textup{attn}}(W)$ defined by
	\[
	\mathcal{W}_{i,j,h} = W_{(h-1)d_h + i,\; j},
	\qquad
	i = 1,\ldots,d_h,\;
	j = 1,\ldots,d_m,\;
	h = 1,\ldots,n_h.
	\]
	The resulting tensor $\mathcal{W} \in \mathbb{R}^{d_h \times d_m \times n_h}$ admits the following interpretation:
	\begin{itemize}
		\item mode-1 ($d_h$): intra-head feature dimension,
		\item mode-2 ($d_m$): input feature dimension,
		\item mode-3 ($n_h$): attention head index.
	\end{itemize}
	Each frontal slice $\mathcal{W}(:,:,h)$ corresponds exactly to the weight submatrix $W_Q^{(h)}$ of the $h$-th attention head.
\end{definition}
\subsubsection{FFN Weight Tensorization}
The feed-forward network consists of an expansion matrix $U_{\textup{in}} \in \mathbb{R}^{d_{ff} \times d_m}$ and a projection matrix $U_{\textup{out}} \in \mathbb{R}^{d_m \times d_{ff}}$, where $d_{ff} = r \cdot d_m$ with expansion ratio $r$ (typically $r = 4$).
\begin{definition}[FFN Input Tensorization Operator]
	\label{def:ffn_in_tensorization}
	Let $U_{\textup{in}} \in \mathbb{R}^{d_{ff} \times d_m}$ with $d_{ff} = r \cdot d_m$.
	The \emph{FFN input tensorization operator}
	\[
	\Phi_{\textup{in}} : \mathbb{R}^{d_{ff} \times d_m} \longrightarrow \mathbb{R}^{d_m \times d_m \times r}
	\]
	is defined by
	\[
	(\mathcal{U}_{\textup{in}})_{i,j,b} = (U_{\textup{in}})_{(b-1)d_m + i,\; j},
	\qquad
	i,j = 1,\ldots,d_m,\;
	b = 1,\ldots,r.
	\]
	Mode-3 indexes the $r$ expansion blocks.
\end{definition}
\begin{definition}[FFN Output Tensorization Operator]
	\label{def:ffn_out_tensorization}
	Let $U_{\textup{out}} \in \mathbb{R}^{d_m \times d_{ff}}$ with $d_{ff} = r \cdot d_m$.
	The \emph{FFN output tensorization operator}
	\[
	\Phi_{\textup{out}} : \mathbb{R}^{d_m \times d_{ff}} \longrightarrow \mathbb{R}^{d_m \times d_m \times r}
	\]
	is defined by
	\[
	(\mathcal{U}_{\textup{out}})_{i,j,b} = (U_{\textup{out}})_{i,\; (b-1)d_m + j},
	\qquad
	i,j = 1,\ldots,d_m,\;
	b = 1,\ldots,r.
	\]
	Mode-3 indexes the $r$ projection blocks.
\end{definition}
\begin{remark}[Unified Mode-3 Semantics]
	Although $U_{\textup{in}}$ and $U_{\textup{out}}$ have transposed shapes, both tensorizations produce tensors of shape $(d_m, d_m, r)$ with mode-3 indexing structural blocks. This ensures that the DCT transform along mode-3 consistently captures inter-block correlations for all FFN weights.
\end{remark}
\begin{proposition}[Properties of Tensorization Operators]
	\label{prop:tensorization_properties}
	All tensorization operators $\Phi \in \{\Phi_{\textup{attn}}, \Phi_{\textup{in}}, \Phi_{\textup{out}}\}$ satisfy:
	\begin{enumerate}
		\item \textbf{Norm Preservation:} $\|\Phi(W)\|_F = \|W\|_F$.
		\item \textbf{Invertibility:} $\Phi^{-1}(\Phi(W)) = W$.
		\item \textbf{Block Decoupling:} Each frontal slice $\mathcal{W}(:,:,k)$ corresponds to a semantically meaningful block (attention head or FFN block).
	\end{enumerate}
\end{proposition}
\begin{figure}[h!]
	\centering
	\begin{tikzpicture}[scale=0.8, transform shape]
		\node[font=\bfseries] at (2, 5.5) {$W_Q \in \mathbb{R}^{d_m \times d_m}$};
		\draw[thick] (0, 0) rectangle (4, 5);
		\fill[red!25] (0, 3.75) rectangle (4, 5);
		\fill[blue!25] (0, 2.5) rectangle (4, 3.75);
		\fill[green!25] (0, 1.25) rectangle (4, 2.5);
		\fill[orange!25] (0, 0) rectangle (4, 1.25);
		\node[font=\scriptsize] at (-0.6, 4.375) {$W_Q^{(1)}$};
		\node[font=\scriptsize] at (-0.6, 3.125) {$W_Q^{(2)}$};
		\node[font=\scriptsize] at (-0.6, 1.875) {$W_Q^{(3)}$};
		\node[font=\scriptsize] at (-0.6, 0.625) {$W_Q^{(4)}$};
		\draw[<->] (0, -0.3) -- (4, -0.3);
		\node[font=\scriptsize] at (2, -0.6) {$d_m$};
		\draw[<->] (4.3, 0) -- (4.3, 5);
		\node[font=\scriptsize, rotate=90] at (4.6, 2.5) {$d_m = n_h \cdot d_h$};
		\draw[->, very thick] (5, 2.5) -- (7, 2.5);
		\node[font=\small, align=center] at (6, 3.2) {$\Phi_{\textup{attn}}$};
		\begin{scope}[shift={(8, 0)}]
			\foreach \h/\col/\off in {1/red!25/0, 2/blue!25/0.5, 3/green!25/1.0, 4/orange!25/1.5} {
				\draw[fill=\col, thick] (\off*0.4, \off*0.3) rectangle (3+\off*0.4, 1.5+\off*0.3);
			}
			\node[font=\bfseries] at (2.5, 4) {$\mathcal{W}_Q \in \mathbb{R}^{d_h \times d_m \times n_h}$};
			\node[font=\scriptsize] at (1.8, -0.3) {$d_m$ (mode-2)};
			\node[font=\scriptsize, rotate=90] at (-0.4, 0.75) {$d_h$ (mode-1)};
			\node[font=\scriptsize] at (4.5, 1.2) {$n_h$ (mode-3)};
			\draw[<-, thick, red!60!black] (0.3, 1.7) -- (-0.5, 2.5);
			\node[font=\scriptsize, red!60!black, align=left] at (-1.2, 3) {Head 1\\slice};
		\end{scope}
	\end{tikzpicture}
	\caption{Attention tensorization $\Phi_{\textup{attn}}$: Weight matrix $W_Q \in \mathbb{R}^{d_m \times d_m}$ is transformed into tensor $\mathcal{W}_Q \in \mathbb{R}^{d_h \times d_m \times n_h}$. Mode-1 indexes intra-head features ($d_h$), mode-2 indexes input features ($d_m$), and mode-3 indexes attention heads ($n_h$). Each frontal slice $\mathcal{W}_Q(:,:,h)$ corresponds to the $h$-th head's weight block.}
	\label{fig:tensorization}
\end{figure}
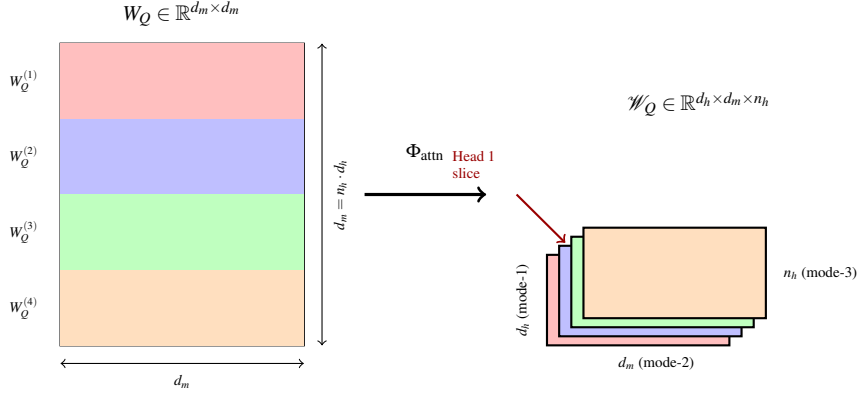
\begin{table}[t]
	\centering
	\caption{Tensorization of GPT-J Weight Matrices}
	\label{tab:tensorization}
	\begin{tabular}{lccc}
		\toprule
		\textbf{Layer Type} & \textbf{Matrix Shape} & \textbf{Tensor Shape} & \textbf{Mode-3}  \\
		\midrule
		Attention (Q,K,V,O) & $(4096, 4096)$ & $(256, 4096, 16)$ & 16 heads  \\
		FFN $U_{\textup{in}}$ & $(16384, 4096)$ & $(4096, 4096, 4)$ & 4 blocks   \\
		FFN $U_{\textup{out}}$ & $(4096, 16384)$ & $(4096, 4096, 4)$ & 4 blocks   \\
		\bottomrule
	\end{tabular}
\end{table}
\begin{remark}[Role of Mode-3 in the c-Product]
	The c-product applies the DCT transform along mode-3. By placing the structural index (heads or blocks) on mode-3, the transform captures correlations \emph{across} these structures:
	\begin{itemize}
		\item For attention: inter-head correlations are captured, exploiting redundancy across the $n_h$ heads.
		\item For FFN: inter-block correlations are captured, exploiting redundancy across the $r$ expansion/projection blocks.
	\end{itemize}
\end{remark}

\begin{table}[t]
\centering
\caption{Feature Comparison: LASER vs TLASER}
\label{tab:comparison}
\begin{tabular}{lcc}
\toprule
\textbf{Property} & \textbf{LASER} & \textbf{TLASER} \\
\midrule
Weight representation & 2D matrix $W \in \R^{d_m \times d_m}$ & 3D tensor $\cal W \in \R^{d_h \times d_m \times n_h}$ \\
Decomposition & Matrix SVD & c-SVD (DCT-based) \\
Arithmetic & Real/Complex & {Real only} \\
Multi-head structure & {Destroyed} & {Preserved} \\
Cross-head correlations & {Ignored} & {Captured via c-product} \\
Energy compaction & Standard & {Superior (DCT)} \\
Partial decomposition & SVD (matrix) & Tensor Bidiag Lanczos (c-product) \\
 \bottomrule
\end{tabular}
\end{table}
\subsection{Tensor dimensions for common architectures }
Modern transformer architectures implement multi-head attention by factorizing the model dimension $d_m$ into $n_h$ attention heads, each operating on a subspace of dimension $d_h$, such that $d_m = n_h \cdot d_h$. Although attention weights are typically stored as flattened matrices $W \in \mathbb{R}^{d_m \times d_m}$, this representation conceals the inherent multi-modal structure induced by multi-head attention. 
To make this structure explicit, we reshape each attention weight matrix into a third-order tensor
\[
\mathcal{W} \in \mathbb{R}^{ d_h \times d_m\times n_h},
\]
where Mode-1 corresponds to the per-head feature dimension ($d_h$), Mode-2 spans the full model dimension ($d_m$), and Mode-3 indexes attention heads ($n_h$). This tensorization is natural across common transformer families such as GPT-J, LLaMA, and Mistral, as summarized in Table~\ref{tab:natural_dimensions}. While the numerical values of $n_h$, $d_h$, and $d_m$ vary between architectures, the same decomposition applies uniformly. 
Viewing attention weights as tensors rather than matrices enables compression methods that respect head-wise and feature-wise correlations. In particular, isolating attention heads as a dedicated tensor mode allows low-rank structure to be exploited across heads without collapsing them into a single matrix dimension. This motivates tensor-based decompositions that operate directly on $\mathcal{W}$ and preserve its multi-modal semantics. 
Figure~\ref{fig:pipeline} illustrates the proposed workflow. A conventional 2D weight matrix is first reshaped into its natural 3D tensor form, after which tensor Golub--Kahan bidiagonalization  is applied to compute only the leading $r$ singular triplets. This avoids the computational cost of a full c-SVD while retaining the dominant interactions across tensor modes, yielding a compressed tensor $\mathcal{W}_r$ that can be mapped back to the original parameter space with minimal loss.

\begin{table}[t]
\centering
\caption{Natural tensor dimensions for attention weights $\cal W \in \R^{n_h \times d_h \times d_m}$}
\label{tab:natural_dimensions}
\begin{tabular}{lcccccc}
\toprule
\textbf{Model} & $d_m$ & $n_h$ & $d_h$ & \textbf{Mode-1} & \textbf{Mode-2} & \textbf{Mode-3} \\
\midrule
GPT-J 6B & 4096 & 16 & 256 & 16 (heads) & 256 (head dim) & 4096 (model) \\
LLaMA-7B & 4096 & 32 & 128 & 32 (heads) & 128 (head dim) & 4096 (model) \\
LLaMA-13B & 5120 & 40 & 128 & 40 (heads) & 128 (head dim) & 5120 (model) \\
Mistral-7B & 4096 & 32 & 128 & 32 (heads) & 128 (head dim) & 4096 (model) \\
\bottomrule
\end{tabular}
\end{table}
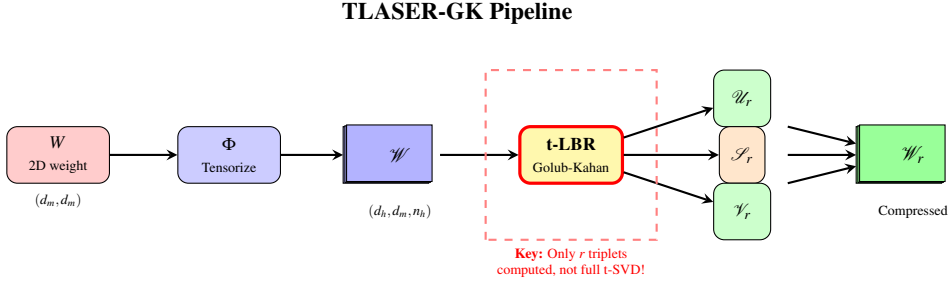
\begin{figure}[t]
\centering
\begin{tikzpicture}[scale=0.75, transform shape,
    block/.style={draw, rounded corners, minimum width=1.8cm, minimum height=1cm, align=center, font=\small},
    arrow/.style={->, thick, >=stealth}
]
    \node[font=\large\bfseries] at (7, 7) {TLASER-GK Pipeline};
    
    \node[block, fill=red!20] (w2d) at (0, 4.5) {$W$\\{\scriptsize 2D weight}};
    
    \node[block, fill=blue!20] (reshape) at (3, 4.5) {
    $\Phi$\\{\scriptsize Tensorize}};
    
    \foreach \i in {0,0.1,0.2} {
        \draw[fill=blue!30] (5+\i*0.3, 4+\i*0.15) rectangle (6.5+\i*0.3, 5+\i*0.15);
    }
    \node at (6, 4.5) {$\mathcal{W}$};
    
    \node[block, fill=yellow!40, very thick, draw=red] (tlbr) at (9, 4.5) {\textbf{t-LBR}\\{\scriptsize Golub-Kahan}};
    
    \node[block, fill=green!20, minimum width=1cm] (U) at (12, 5.5) {$\mathcal{U}_r$};
    \node[block, fill=orange!20, minimum width=0.8cm] (S) at (12, 4.5) {$\mathcal{S}_r$};
    \node[block, fill=green!20, minimum width=1cm] (V) at (12, 3.5) {$\mathcal{V}_r$};
    
    \foreach \i in {0,0.1,0.2} {
        \draw[fill=green!40] (14+\i*0.3, 4+\i*0.15) rectangle (15.5+\i*0.3, 5+\i*0.15);
    }
    \node at (15, 4.5) {$\mathcal{W}_r$};
    
    \draw[arrow] (w2d) -- (reshape);
    \draw[arrow] (reshape) -- (5, 4.5);
    \draw[arrow] (6.7, 4.5) -- (tlbr);
    \draw[arrow] (tlbr) -- (U);
    \draw[arrow] (tlbr) -- (S);
    \draw[arrow] (tlbr) -- (V);
    \draw[arrow] (12.8, 5) -- (14, 4.7);
    \draw[arrow] (12.8, 4.5) -- (14, 4.5);
    \draw[arrow] (12.8, 4) -- (14, 4.3);
    
    \node[font=\scriptsize] at (0, 3.7) {$(d_m, d_m)$};
    \node[font=\scriptsize] at (6, 3.5) {$( d_h, d_m,n_h)$};
    \node[font=\scriptsize] at (15, 3.5) {Compressed};
    
    \draw[dashed, red!50, thick] (7.5, 3) rectangle (10.5, 6);
    \node[font=\scriptsize, red, align=center] at (9, 2.6) {\textbf{Key:} Only $r$ triplets\\computed, not full t-SVD!};
     
\end{tikzpicture}
\caption{TLASER-GK pipeline: reshape 2D weights to 3D tensors, apply t-LBR (Golub-Kahan bidiagonalization) to compute only the $r$ needed singular triplets, reconstruct compressed weights.}
\label{fig:pipeline}
\end{figure}

\medskip
 \subsection{Low-Rank Approximation of Layer Weights}

Consider a pretrained neural network with layers indexed by $\ell = 1, \dots, L$. Let
\[
{\mathcal W}^{(\ell)} \in \mathbb{R}^{m_\ell \times n_\ell \times p_\ell}
\]
denote a weight tensor associated with layer $\ell$. For a target rank
$r_\ell \ll \min(m_\ell,n_\ell)$, LASER replaces ${\mathcal W}^{(\ell)}$ by its truncated singular
value decomposition
\[
{\mathcal W}^{(\ell)} \approx \widetilde{\mathcal{W}}_{r_\ell}^{(\ell)}=  {\mathcal U}_{r_\ell}^{(\ell)} *_c {\mathcal S}_{r_\ell}^{(\ell)} *_c{\mathcal V}_{r_\ell}^{(\ell)\top},
\]
where
\[
{\mathcal U}_{r_\ell}^{(\ell)} \in \mathbb{R}^{m_\ell \times r_\ell \times p_\ell}, \quad
{\mathcal S}_{r_\ell}^{(\ell)} \in \mathbb{R}^{r_\ell \times r_\ell \times p_\ell}, \quad
{\mathcal V}_{r_\ell}^{(\ell)} \in \mathbb{R}^{n_\ell \times r_\ell \times p_\ell}.
\]


Tensor LASER introduces a layer-wise selection mechanism to determine which layers are
compressed. Define a binary selection variable
\[
\delta_\ell =
\begin{cases}
1, & \text{if layer $\ell$ is compressed}, \\
0, & \text{otherwise}.
\end{cases}
\]
The effective weight tensor employed during inference or training in layer $\ell$ is therefore defined as
\[
\widetilde{\mathcal{W}}_{r_\ell}^{(\ell)} =
\begin{cases}
{\mathcal U}_{r_\ell}^{(\ell)} *_c {\mathcal S}^{(\ell)}_{r_\ell} *_c{\mathcal V}_{r_\ell}^{(\ell)\top} & \;  if \; \delta_\ell = 1, \\
\mathcal{W}^{(\ell)}, & \; if \; \delta_\ell = 0.
\end{cases}
\]
Layer selection may be guided by singular value decay, sensitivity analysis, or empirical
heuristics, such as compressing intermediate layers more aggressively than early or late
layers.
\\\
Tensor LASER introduces a \emph{layer-wise selection mechanism} that determines which layers of the network are subjected to low-rank tensor compression. Rather than compressing all layers uniformly, this strategy enables selective rank reduction, thereby preserving performance-critical components while reducing redundancy in overparameterized layers.
Let $\ell \in \{1,\dots,L\}$ index the layers of the model. We introduce a binary selection variable
\begin{equation}
\delta_\ell =
\begin{cases}
1, & \text{if layer } \ell \text{ is selected for compression},\\
0, & \text{otherwise}.
\end{cases}
\end{equation}
This variable explicitly encodes whether low-rank tensor approximation is applied to the weight tensor of layer $\ell$. 
The choice of $\delta_\ell$ can be guided by several complementary criteria. A common approach relies on the decay behavior of the singular values (or singular tubes) of $\mathcal{W}^{(\ell)}$, where a rapid decay indicates a high degree of redundancy and thus a strong candidate for compression. Alternatively, sensitivity analysis may be employed to measure the impact of perturbations in $\mathcal{W}^{(\ell)}$ on the task loss, allowing layers with low sensitivity to be compressed more aggressively.
In practice, empirical heuristics are often effective. For instance, intermediate layers in deep Transformer architectures typically exhibit greater redundancy and can tolerate stronger rank reduction, whereas early layers (which encode low-level representations) and late layers (which are closely tied to task-specific outputs) are compressed more conservatively or left uncompressed.\\
By combining tensor-based low-rank approximation with explicit layer selection, Tensor LASER achieves a favorable trade-off between compression rate and model accuracy. This selective strategy reduces computational and memory costs while maintaining the expressive capacity of the most critical layers, making it particularly suitable for large-scale Transformer and large language models.

\subsection{Rank Selection Strategy}
\subsubsection{Rank allocation}
A central component of the Tensor LASER framework is the determination of suitable
truncation ranks for constructing accurate low-rank tensor approximations of the
model parameters. In large language models, each layer $\ell$ is associated with a
high-dimensional weight tensor $\mathcal{W}^{(\ell)}$, whose size and redundancy
vary significantly across layers. Applying the same rank reduction uniformly across
all layers is therefore suboptimal and may lead to unnecessary performance degradation.
\\
To address this issue, Tensor LASER performs rank selection at the layer level.
Specifically, given a weight tensor $\mathcal{W}^{(\ell)}$, its tensor singular value
decomposition under the $c$-product is written as
\[
\mathcal{W}^{(\ell)} =
\mathcal{U}^{(\ell)} *_c
\mathcal{S}^{(\ell)} *_c
\mathcal{V}^{(\ell)\top},
\]
where $\mathcal{U}^{(\ell)}$ and $\mathcal{V}^{(\ell)}$ are $c$-orthogonal tensors,
and $\mathcal{S}^{(\ell)}$ is an $f$-diagonal tensor whose diagonal fibers
(\emph{singular tubes}) encode the spectral content of $\mathcal{W}^{(\ell)}$.
These singular tubes are conventionally ordered according to their $\ell_2$ norms
in non-increasing order, such that the leading tubes capture the most significant
components of the weight tensor.
\\
The objective of rank selection is to identify the smallest truncation rank
$r_\ell$ for which the truncated approximation
\[
\mathcal{W}^{(\ell)}_{r_\ell} =
\mathcal{U}^{(\ell)}_{r_\ell} *_c
\mathcal{S}^{(\ell)}_{r_\ell} *_c
\mathcal{V}^{(\ell)\top}_{r_\ell}
\]
retains the dominant information of the original tensor while discarding
redundant or low-energy components. In practice, this amounts to striking a
balance between approximation accuracy and parameter efficiency: increasing
$r_\ell$ improves fidelity but incurs higher storage and computational costs,
whereas smaller ranks yield stronger compression at the risk of performance loss.

By tailoring the truncation rank to each individual layer, Tensor LASER exploits
the heterogeneous spectral structure of modern deep architectures. Layers with
rapid singular value decay can be aggressively compressed, while layers that are
more sensitive to rank reduction are preserved with higher ranks. This adaptive
rank selection mechanism is a key factor enabling Tensor LASER to achieve
substantial model compression while maintaining task-level performance.

\subsubsection{Energy-Based Rank Selection.}

A commonly used criterion is based on the cumulative energy captured by the leading singular tubes. Let
\[
\bm{s}^{(\ell)}_i = \mathcal{S}^{(\ell)}(i,i,:) , \qquad i=1,\ldots,\min(m,n),
\]
denote the singular tubes of $\mathcal{W}^{(\ell)}$. The rank $r_\ell$ is chosen as the smallest integer satisfying
\begin{equation}
\displaystyle\frac{\displaystyle \sum_{i=1}^{r_\ell} \| \bm{s}^{(\ell)}_i \|_2^2}
     {\displaystyle \sum_{i=1}^{\min(m,n)} \| \bm{s}^{(\ell)}_i \|_2^2}
\ge \tau,
\label{eq:energy_rank}
\end{equation}
where $\tau \in (0,1)$ is a user-defined energy retention threshold, typically chosen in the range $[0.9, 0.99]$.

\subsubsection{Layer-Adaptive Rank Allocation.}
In deep architectures, different layers exhibit varying degrees of redundancy. Tensor LASER therefore allows the rank $r_\ell$ to be chosen adaptively across layers. In practice, larger ranks are assigned to early and late layers, while more aggressive rank reduction is applied to intermediate layers, which typically admit faster singular value decay.
\\
Beyond purely spectral criteria, rank selection can also be guided by task-level performance. Let $\mathcal{L}(\cdot)$ denote the task loss (e.g., cross-entropy). The rank $r_\ell$ may be selected by solving
\[
r_\ell^\star = \arg\min_{r_\ell}
\mathcal{L}\!\left(\widehat{\mathcal{W}}^{(\ell)}_{r_\ell}\right)
\quad
\text{subject to }
\left\| \mathcal{W}^{(\ell)} - \mathcal{W}^{(\ell)}_{r_\ell} \right\|_F
\le \varepsilon_\ell,
\]
which balances compression against downstream performance.
\\
The proposed rank selection strategies provide flexible mechanisms to trade off accuracy, compression ratio, and computational efficiency. When combined with layer-selective compression, rank selection becomes a powerful tool for scaling Tensor LASER to large Transformer and large language models.

\medskip
A key advantage of Tensor LASER lies in its ability to exploit the intrinsic multi-dimensional structure of Transformer weights through tensor representations.
By operating directly on tensorized parameters and employing transform-based tensor products, Tensor LASER captures correlations across attention heads and feature
dimensions that are overlooked by matrix-based compression methods. This structured low-rank approximation leads to higher compression efficiency for a given accuracy level, while preserving the architectural semantics of multi-head attention.
Moreover, Tensor LASER is inherently training-free, requiring neither fine-tuning nor additional data, which makes it particularly well suited for large-scale models and resource-constrained deployment settings.

\begin{algorithm}[t]
\caption{Tensor LASER algorithm (TLASER)}
\label{alg:tlaser_complete}

\textbf{Input:}
Model with layers $\{\mathcal{W}^{(\ell)}\}_{\ell=1}^L$, energy threshold $\tau$, layer selection indicators $\{\delta_\ell\}$.

\textbf{Output:}
Compressed model.

\medskip
\textbf{Procedure:}
\begin{enumerate}
    \item For each layer $\ell = 1, \ldots, L$:
    \begin{enumerate}
        \item If $\delta_\ell = 1$:
        \begin{enumerate}
            \item {Weight tensor construction.}  
            Let $W^{(\ell)}$ be the weight matrix of layer $\ell$ and reshape it as
            \[
            \mathcal{W}^{(\ell)} = \text{reshape}(W^{(\ell)}, [d_h, d_m, n_h]).
            \]
            \item { c-SVD decomposition.}  
            Apply the discrete cosine transform along the third mode:
            \[
            \widehat{\mathcal{W}}^{(\ell)} = \text{DCT}(\mathcal{W}^{(\ell)}, 3).
            \]
            For $i = 1, \ldots, p$, compute the matrix SVD
            \[
            \widehat{W}^{(\ell,i)} =
            \widehat{U}^{(i)} \widehat{S}^{(i)} (\widehat{V}^{(i)})^T .
            \]
            \item {Rank determination.}
            Compute the singular tube norms $\|\bm{s}_i^{(\ell)}\|$ for
            $i = 1, \ldots, \min(d_h, d_m)$, and select
            \[
            r_\ell =
            \min \left\{
            r :
            \frac{\sum_{i=1}^r \|\bm{s}_i^{(\ell)}\|}
                 {\sum_i \|\bm{s}_i^{(\ell)}\|}
            \geq \tau
            \right\}.
            \]
            \item {Truncation.}  
            For $i = 1, \ldots, p$, construct the truncated slices
            \[
            \widehat{W}_r^{(\ell,i)} =
            \widehat{U}^{(i)}_{:,:r_\ell}
            \widehat{S}^{(i)}_{:r_\ell,:r_\ell}
            (\widehat{V}^{(i)}_{:,:r_\ell})^T .
            \]
            \item {Inverse transform and update.}  
            Apply the inverse discrete cosine transform:
            \[
            \mathcal{W}_r^{(\ell)} =
            \text{IDCT}(\widehat{\mathcal{W}}_r^{(\ell)}, 3).
            \]
        \end{enumerate}
    \end{enumerate}
\end{enumerate}
\end{algorithm}

 \newpage 
\subsection{Experimental results}
\label{sec:protocol}

We conducted experiments on GPT-J-6B~\cite{wang2021gptj}, a 6 billion parameter autoregressive language model with 28 transformer layers. The model is characterized by: model dimension $d_{\text{model}} = 4096$, number of attention heads $n_h = 16$, head dimension $d_h = 256$, and FFN intermediate dimension $d_{\text{ff}} = 16384$.  We tensorized weight matrices according to their inherent architectural structure. Table~\ref{tab:tensorization} summarizes our tensorization scheme.



\nd We evaluated our method on TruthfulQA~\cite{Lin2021}, a benchmark measuring the extent to which language models generate truthful answers. The evaluation uses the pointwise protocol where each (question, answer) pair is converted to a binary classification task.
The processed dataset contains 5,882 samples. We evaluate on $N=500$ samples with layer $\ell = 27$ and rank ratio $\rho = 0.70$. 
Following~\cite{Sharma2023}, we reported two metrics:
\begin{itemize}
\item  {\bf Classification Accuracy (Acc)}: A response is correct if the model assigns higher probability to the correct label than to any other candidate:
\begin{equation}
    \text{Acc} = \frac{1}{N} \sum_{i=1}^{N} \mathbf{1}\left[ \arg\max_{y \in \{\text{true}, \text{false}\}} P(y | x_i) = y_i^* \right]
\end{equation}

\item {\bf Loss} The log-loss on held-out data:
\begin{equation}
    \text{Loss} = -\frac{1}{N} \sum_{i=1}^{N} \log P(y_i^* | x_i)
\end{equation}
Lower loss indicates higher model confidence in correct predictions.
\end{itemize}

\nd Table~\ref{tab:laser_paper} presents the original LASER results from~\cite{Sharma2023}.  The optimal LASER configuration for GPT-J on TruthfulQA is $[U_{\text{in}}, \ell=7, \rho=0.80]$. Table~\ref{tab:single_layer} presents results for single-layer interventions where TLASER demonstrates improved performance.

\begin{table}[!t]
\centering
\caption{LASER Results from~\cite{Sharma2023} on TruthfulQA}
\label{tab:laser_paper}
\begin{tabular}{lcccc}
\toprule
\textbf{Model} & \textbf{Acc} & \textbf{LASER Acc} & \textbf{Loss} & \textbf{LASER Loss} \\
\midrule
GPT-J & 54.9\% & 55.6\% & 1.02 & 1.01 \\
\bottomrule
\end{tabular}
\end{table}



\begin{table}[!t]
\centering
\caption{Single-Layer Interventions: TLASER vs LASER (GPT-J, Layer 27, $\rho=0.70$, $N=500$)}
\label{tab:single_layer}
\begin{tabular}{lcccc}
\toprule
\multirow{2}{*}{\textbf{Intervention}} & \multicolumn{2}{c}{\textbf{LASER}} & \multicolumn{2}{c}{\textbf{TLASER}} \\
\cmidrule(lr){2-3} \cmidrule(lr){4-5}
& Acc (\%) & Loss & Acc (\%) & Loss \\
\midrule
$U_{\text{in}}$ only & 56.00 & 0.977 & {56.40} & 1.008 \\
$U_{\text{out}}$ only & 56.00 & 1.021 & \textbf{56.40} & {1.012} \\
\bottomrule
\end{tabular}
\end{table}

\begin{itemize}
    \item For $U_{\text{in}}$ intervention: TLASER achieves \textbf{+0.40\%} accuracy improvement (56.40\% vs 56.00\%).
    \item For $U_{\text{out}}$ intervention: TLASER achieves \textbf{+0.40\%} accuracy improvement and \textbf{lower loss} (1.012 vs 1.021).
\end{itemize}

\nd Table~\ref{tab:config} compares our experimental setup with the optimal LASER configuration.

\begin{table}[!t]
\centering
\caption{Configuration Comparison with LASER Paper}
\label{tab:config}
\begin{tabular}{lcc}
\toprule
\textbf{Parameter} & \textbf{LASER Paper} & \textbf{Our Experiments} \\
\midrule
Target Layer & $\ell = 7$ (early) & $\ell = 27$ (late) \\
Rank Ratio & $\rho = 0.80$ & $\rho = 0.70$ \\
Target Matrix & $U_{\text{in}}$ & $U_{\text{in}}$, $U_{\text{out}}$, both \\
\bottomrule
\end{tabular}
\end{table}

\nd The layer discrepancy ($\ell=7$ vs $\ell=27$) suggests that TLASER may have different optimal intervention points than LASER, warranting future layer-sweep experiments.


\begin{table}[!t]
\centering
\caption{Summary: LASER vs TLASER}
\label{tab:summary}
\begin{tabular}{lcc}
\toprule
\textbf{Criterion} & \textbf{LASER} & \textbf{TLASER} \\
\midrule
Reconstruction Error & 0.31--0.35 & {0.13--0.14} \\
Error Improvement & 1.0$\times$ & {2.5$\times$} \\
\midrule
Accuracy (single-layer) & 56.00\% & {56.40\%} \\
\midrule
Structure Preservation & Destroyed & {Preserved} \\
Inter-block Correlations & Lost & {Captured} \\
\midrule
\bottomrule
\end{tabular}
\end{table}

\subsection{Comments on the Experimental Protocol.}
The experimental protocol is designed to isolate the effect of the rank-reduction method while controlling for architectural, data, and evaluation factors. GPT-J-6B provides a challenging large-scale setting with strong structural regularities, particularly in its multi-head attention and block-structured feed-forward layers. By keeping the model architecture fixed and intervening on a single late transformer layer ($\ell = 27$), the protocol focuses on localized weight modifications that directly affect high-level semantic representations.
\\
The tensorization strategy in Table~\ref{tab:tensorization} explicitly aligns the decomposition with the architectural organization of the model. Attention matrices are reshaped along the head dimension, while FFN matrices are partitioned into blocks reflecting the expansion and contraction structure. This design choice ensures that TLASER operates on semantically meaningful modes rather than arbitrary reshapes, enabling a fair comparison with LASER under identical rank ratios.
\\
The use of TruthfulQA follows prior work and provides a sensitive benchmark for detecting subtle changes in model behavior. The pointwise evaluation protocol converts generation into a controlled binary classification task, reducing variance due to decoding strategies and making accuracy and loss directly comparable across interventions. Evaluating on a fixed subset of $N=500$ samples allows efficient experimentation while remaining consistent with the LASER evaluation methodology.
\\
Table~\ref{tab:laser_paper} reports the original LASER results, where the optimal configuration targets an early FFN layer ($\ell=7$) with a relatively high rank ratio. In contrast, our experiments intentionally explore a different regime, focusing on a later layer with stronger semantic abstraction and a more aggressive compression level ($\rho=0.70$). This shift allows us to assess whether tensor-based decompositions behave differently from matrix SVD when applied to deeper representations.
\\
The results in Table~\ref{tab:single_layer} show that TLASER consistently improves accuracy over LASER when a single FFN matrix is modified. The $+0.40\%$ gain observed for both $U_{\text{in}}$ and $U_{\text{out}}$ under identical settings suggests that preserving structured correlations within a single layer leads to less disruptive edits. The improvement is particularly meaningful given that both methods apply the same rank ratio and intervention depth.
\\
The loss behavior further highlights functional differences between the two matrices. For $U_{\text{out}}$, TLASER improves both accuracy and loss, indicating more confident predictions. For $U_{\text{in}}$, the slight increase in loss alongside higher accuracy suggests that structure preservation may sharpen decision boundaries without uniformly improving likelihood estimates.
\\
When both FFN matrices are modified simultaneously (Table~\ref{tab:multi_layer}), LASER exhibits a small advantage in both accuracy and loss. This reversal indicates that the benefits of structure-preserving tensor decompositions do not necessarily compose across consecutive layers. Since the FFN implements a nonlinear transformation between $U_{\text{in}}$ and $U_{\text{out}}$, preserving correlations in both matrices may introduce interaction effects that are less favorable after the nonlinearity. In this setting, the stronger decorrelation induced by matrix SVD may act as a stabilizing factor.
\\

\nd The summary in Table~\ref{tab:summary} highlights a clear interaction between intervention scope and decomposition method. TLASER is consistently advantageous for single-layer interventions, while LASER remains competitive—and slightly superior—for joint multi-layer modifications. Importantly, the magnitude of the differences remains symmetric ($\pm 0.40\%$), indicating comparable overall performance with complementary strengths.
\\
Given the evaluation size, these results should be interpreted as directional trends rather than statistically significant differences. Nevertheless, they suggest that TLASER is particularly well suited for targeted, fine-grained interventions where preserving internal structure is beneficial, whereas LASER may be preferable for broader, multi-layer rank reductions.\\
\vskip0.2cm
\nd Our experiments on TruthfulQA demonstrate that 
\begin{itemize}
    \item {TLASER improves single-layer interventions:} For both $U_{\text{in}}$ and $U_{\text{out}}$ interventions individually, it  achieves +0.40\% accuracy over LASER.
    \item {Interaction effect for multi-layer interventions:} When both MLP matrices are modified, LASER shows a +0.40\% advantage, suggesting compositional effects in the FFN pathway.
    \item {Method selection depends on intervention scope:} TLASER is preferable for targeted single-layer edits, while LASER may be more suitable for broader multi-layer modifications.
\end{itemize}

\section{Conclusion}

In this work, we introduced Tensor Layer-Selective Rank Reduction (TLASER), a unified
tensor-based framework for efficient compression of large Transformer models and large language models. TLASER extends matrix-based layer-selective rank reduction techniques to
a principled tensor setting by leveraging the $\mathcal{L}$-product framework, with a particular emphasis on the DCT-based c-product. This approach enables structured, real-valued tensor algebra that preserves the intrinsic multi-head organization of
Transformer weights while capturing cross-head correlations that are ignored by classical
matrix methods.
\\
By formulating attention and feed-forward weights as third-order tensors, TLASER exploits the natural multi-dimensional structure of modern architectures. The use of the c-SVD
provides a direct generalization of matrix spectral analysis, while tensor Lanczos bidiagonalization enables scalable computation of only the dominant singular triplets,
making the method suitable for large-scale models where full decompositions are computationally prohibitive. The resulting low-rank tensor approximations offer substantial reductions in parameter count, memory footprint, and computational complexity, without requiring retraining or access to additional data.


\begin{thebibliography}{99}

\bibitem{Achiam2023}
J. Achiam, S. Adler, S. Agarwal, L. Ahmad, I. Akkaya, F. L. Aleman, D. Almeida, J. Altenschmidt, S. Altman, S. Anadkat, et al., {GPT-4 technical report},
arXiv preprint arXiv:2303.08774, 2023.

\bibitem{Aeron2015}
S. Aeron, E. Kernfeld, and M. Kilmer, Tensor-tensor products with invertible linear transforms, Linear Algebra and Its Applications, 485, 545--570, 2015.

\bibitem{Ash2023}
J.~T.~Ash, P.~Sharma, and D.~Misra,
{Rank Reduction for Efficient Transformer Compression}, arXiv preprint arXiv:2305.XXXX, 2023.

\bibitem{Brown2020}
T. Brown, B. Mann, N. Ryder, M. Subbiah, J. D. Kaplan, P. Dhariwal, 
A. Neelakantan, P. Shyam, G. Sastry, A. Askell, et al., {Language models are few-shot learners},
Advances in Neural Information Processing Systems, 33, 1877--1901, 2020.

\bibitem{dearteaga2019bias}
M. De-Arteaga, Alexey Romanov, Hanna Wallach, et al., Bias in Bios: A Case Study of Semantic Representation Bias in a High-Stakes Setting.
 In \emph{FAT*}, 2019.

\bibitem{Dufrenois2023}
F. Dufrenois, A. El Ichiband  K. Jbilou, 
Multilinear discriminant analysis using tensor-tensor products, Journal of Mathematical Modeling,  11(1), 83--101, 2023.

\bibitem{Elhachimi2024}
A. El Hachimi, K. Jbilou, A.  Ratnani, and L. Reichel,
A Tensor Bidiagonalization Method for Higher-Order Singular Value Decomposition with Applications, Numerical Linear Algebra with Applications, 31(2), e2530, 2024.

 \bibitem{Elhachimi2023}
 A. El Hachimi, K. Jbilou, A. Ratnani, and L. Reichel, Spectral computation with  third-order tensors using the t-product, Applied  Numerical  Mathematics, 193,1--23, 2023.

\bibitem{Hached2021}
M. Hached, K. Jbilou, C. Koukouvinos, M. Mitrouli, A multidimensional principal component analysis via the C-product Golub–Kahan–SVD for classification and face recognition, Mathematics,  9(11), 1249, 2021.

\bibitem{Hitchcock1927}
F. L. Hitchcock,
{The expression of a tensor or a polyadic as a sum of products}, Journal of Mathematics and Physics, 6(1-4), 164–189 (1927).


\bibitem{Kilmer2011}
M. E. Kilmer, and C. D. Martin, Factorization strategies for third-order tensors, Linear  Algebra and Applications, 435, 641--658, 2011.

\bibitem{Kolda2009}
T. G. Kolda and B. W. Bader, Tensor decompositions and applications, SIAM Review,  51, 455--500,  2009.

\bibitem{Lin2021}
S.  Lin, Jacob Hilton, and Owain Evans.
\newblock TruthfulQA: Measuring How Models Mimic Human Falsehoods. In \emph{ACL}, 2022.

\bibitem{Luo2024}
Y. Luo, H. Patel, Y. Fu, D. Ahn, J. Chen, Y. Dong, and E. E. Papalexakis, TRAWL: Tensor Reduced and Approximated Weights for Large Language Models, arXiv preprint arXiv:2406.17261, 2024.

\bibitem{Newman2024}
E. Newman, L. Horesh, H. Avron, and M. E. Kilmer,
 Stable tensor neural networks for efficient deep learning, Frontiers in Big Data, 7, 1363978, 2024.
doi:10.3389/fdata.2024.1363978

 \bibitem{Oseledets2011}
 I. V. Oseledets, Tensor-Train Decomposition,
 SIAM Journal on Scientific Computing, 33(5), 2295–2317, 2011.

\bibitem{Sharma2023}
P.~Sharma, J.~T.~Ash, and D.~Misra,
{The Truth Is in There: Improving Reasoning in Language Models with Layer-Selective Rank Reduction},
 in Proceedings of the Twelfth International Conference on Learning Representations (ICLR), 2024.

\bibitem{Touvron2023}
H. Touvron, T. Lavril, G. Izacard, X. Martinet, M.-A. Lachaux,  T. Lacroix, B. Rozi`ere, N. Goyal, E. Hambro, F. Azhar, et al.,
LLaMA: Open and efficient foundation language models,
arXiv preprint arXiv:2302.13971, 2023.

\bibitem{Tucker1966}
L. R. Tucker, {Some mathematical notes on three-mode factor analysis}, Psychometrika, 31(3), 279--311, 1966.

\bibitem{Vaswani2017}
A. Vaswani, N. Shazeer, N. Parmar, J. Uszkoreit, L. Jones, A. N. Gomez, L. u. Kaiser, and I. Polosukhin, Attention is All you Need, in Advances in Neural Information Processing Systems, vol. 30, 2017.


\bibitem{Wang2019}
Q. Wang, B. Li, T. Xiao, J. Zhu, C. Li, D. F. Wong, and L. S. Chao, Learning Deep Transformer Models for Machine Translation, in Proceedings of the 57th Annual Meeting of the Association for Computational Linguistics, pp. 1810–1822, July 2019.

\bibitem{Xia2024}
W.~Xia, R.~Zhang, Z.~Li, T.~Chen, and Z.~Wang,
 {LASER: Layer-Selective Rank Reduction for Compressing Large Language Models}, {arXiv preprint arXiv:2403.01242}, 2024.


\bibitem{Xu2023}
M. Xu, Y. L. Xu, and D. P. Mandic,
TensorGPT: Efficient compression of the embedding layer in LLMs based on the tensor-train decomposition, arXiv preprint arXiv:2307.00526, 2023.

\bibitem{Yu2020}
M.~Yu, X.~Wang, W.~Shi, and B.~Liu,
{Compressing Transformers: Features Are Low-Rank but Weights Are Not}, in Proceedings of the AAAI Conference on Artificial Intelligence, 2020.



\bibitem{wang2021gptj}
B. Wang and A.  Komatsuzaki.
\newblock GPT-J-6B: A 6 Billion Parameter Autoregressive Language Model.  \url{https://github.com/kingoflolz/mesh-transformer-jax}, 2021.







\end{thebibliography}
\end{document}